\pdfoutput=1

\documentclass[11pt]{article}

\usepackage{acl}

\usepackage{times}
\usepackage{latexsym}

\usepackage[T1]{fontenc}

\usepackage[utf8]{inputenc}

\usepackage{microtype}

\usepackage{graphicx}
\usepackage{tabularx}
\usepackage{array}
\usepackage{multirow}
\usepackage{hyperref}
\title{Named Entity Recognition in Indian court judgments}

\author{Prathamesh Kalamkar$^{1,2,*}$, Astha Agarwal$^{1,2,*}$, Aman Tiwari$^{1,2,*}$, Smita Gupta$^{3,}$\thanks{* Authors contributed equally}, \\ {\bf \large Saurabh Karn$^{3,*}$, Vivek Raghavan$^{1}$} \\
$^{1}$EkStep Foundation,  $^{2}$Thoughtworks Technologies India Pvt Ltd., $^{3}$Agami \\
\{prathamk, aman.tiwari, astha.agarwal\}@thoughtworks.com,\\ \{smita, saurabh\}@agami.in, vivek@ekstep.org}

\begin{document}
\maketitle
\newcommand\PK[1]{\textbf{\textcolor{red}{(#1)$_{Prathamesh}$}}}
\newcommand\SG[1]{\textbf{\textcolor{blue}{(#1)$_{Smita}$}}}
\newcommand\AT[1]{\textbf{\textcolor{green}{(#1)$_{Aman}$}}}
\newcommand\AG[1]{\textbf{\textcolor{orange}{(#1)$_{Astha}$}}}

\begin{abstract}
Identification of named entities from legal texts is an essential building block for developing other legal Artificial Intelligence applications. Named Entities in legal texts are slightly different and more fine-grained than commonly used named entities like Person, Organization, Location  etc. In this paper, we introduce a new corpus of 46545 annotated legal named entities mapped to 14 legal entity types. The Baseline model for extracting legal named entities from judgment text is also developed. We publish the training, dev data and trained baseline model \url{https://github.com/Legal-NLP-EkStep/legal_NER}.
\end{abstract}
\section{Introduction}

Artificial Intelligence has the potential to increase access to justice and make various legal processes more efficient \cite{zhong2020does}. Populous countries such as India have a problem with high  case pendency. As of March 2022, over 47 million cases are pending in Indian courts\footnote{\url{https://www.livelaw.in/pdf_upload/au595-426886.pdf}}. Hence, it becomes imperative to use AI to reduce the strain on the judicial system and reduce pendency. 
For developing legal AI applications, it is essential to have access to judicial data and open-source foundational AI building blocks like Named Entity Recognition (NER). A lot of Indian legal data is publicly available thanks to open data initiatives like National Judicial Data Grid (NJDG) and the Crime and Criminal Tracking Network and System (CCTNS).

NJDG provides non-exhaustive metadata of Indian court judgements like the names of petitioners, respondents, lawyers, judges, date, court etc.
Extracting these entities from judgment text makes the information extraction exhaustive and reduces errors like misspellings compared to NJDG metadata. Helpful information like precedents and statutes are also not written in the NJDG metadata. Hence it is essential to extract from court judgment texts rather than just relying on the published NJDG metadata. Extracting named entities from the text also paves the foundation for more tasks like relation extraction, coreference resolution, knowledge graph creation etc. 

\begin{figure*}[ht]
\begin{center}
\includegraphics[scale=0.5]{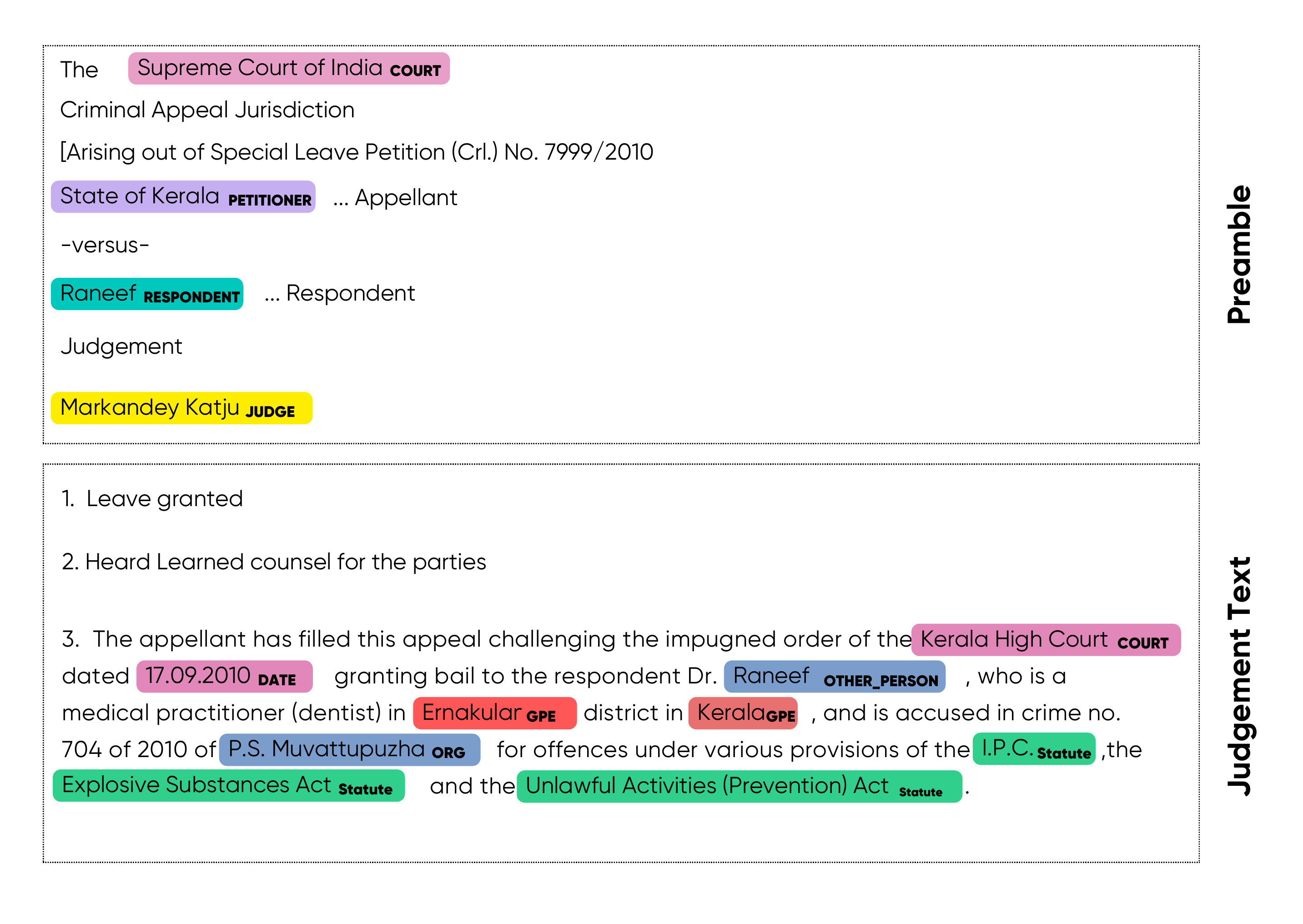} 
\caption{Legal Named Entities in a court judgment}
\label{fig:NER_example}
\end{center}
\end{figure*}

In this paper, we have created a corpus of annotated judgment texts with 14 legal entities (details in \S \ref{sec:NER_dataset}). An example of annotated entities is shown in Figure \ref{fig:NER_example}.   

We make the following contributions in this paper
\begin{itemize}
\item We create a corpus of 14444 Indian court judgment sentences and 2126 judgment preambles annotated with 14 legal named entities.
\item We develop a transformer-based legal NER baseline model 
\item We  create rule-based post-processing, which captures the document level context and coreference resolution for certain entities 
\item A representative sample of Indian high court and supreme court judgments having 11970 judgments across 29 Indian courts
\end{itemize}

\section{Related Work}
Named Entity Recognition (NER) is widely studied in literature ranging from statistical models \cite{borthwick1998nyu},\cite{bikel1999algorithm},\cite{mccallum2003early} to state-of-the-art deep neural nets \cite{li2020survey}. The task complexity is also evolved over time from flat named entities to nested entities, from monolingual to multilingual NER.

Legal domain-specific entities are often used for more meaningful information extraction from legal texts. Pioneering work in legal NER by \cite{dozier2010named} developed named entity recognition \& resolution system on US legal texts using 5 legal named entities (judges, attorneys, companies, jurisdictions, and courts). \cite{cardellino2017low} created  Named Entity Recognizer, Classifier and Linker by mapping LKIF ontology to YAGO ontology using Wikipedia data and various levels of abstraction of the legal ontology. 
Since the legal vocabulary and style of writing of legal text varies by language and geography, it is often necessary to create separate datasets and models. \cite{glaser2018named} compared GermaNER \cite{benikova2015c} and DBpedia Spotlight \cite{mendes2011dbpedia,daiber2013improving} NER systems on German legal contracts. \cite{leitner2020dataset} created a German NER dataset with 19 fine-grained semantic classes. 
\cite{puais2021named} created a Romanian legal corpus called Legal NERo, which has 370 documents annotated with five entity classes and used legal domain word embeddings to build the NER system. 
\cite{luz2018lener} created a corpus of legal documents from several Brazilian Courts called LeNER-Br, which is annotated with six entity classes.  \cite{angelidis2018named} created Named Entity Recognizer and Linker for Greek legislation with 254 annotated pieces of legislation. \cite{chalkidis2021neural} extracted contract elements extraction using LSTM encoders.
NER using contextual dictionaries was applied to the French legal corpus of 94 court judgments with four entity classes by \cite{barriere2019may}.
As a part of Lynx, project \cite{schneider2020orchestrating}, a set of services, including NER, were developed to help create a legal domain knowledge graph and its use for the semantic analysis of legal documents.

Transition-based parsing for NER was proposed by \cite{lample2016neural} using stacked LSTM. NER task can be treated as graph-based dependency parsing \cite{yu2020named} to provide a global view of the input using biaffine model. Recent advances in span representation have shown promising results for Named Entity Recognition \cite{ouchi2020instance}. Span pretraining methods \cite{joshi2020spanbert} improve the span representation for pre-trained language models via span-level pretraining tasks. Infusing external knowledge for entity representation and linking \cite{yamada2020luke}, \cite{wang2021kepler} helps to better represent the knowledge in legal texts.   \cite{ye2022packed} considered interrelation between spans by considering the neighbouring spans integrally to better model the entity boundary information.

Recently a lot of work has been done in the legal AI field in the Indian context. Structuring the court judgments \cite{kalamkar-EtAl:2022:LREC}, legal statute identification \cite{paul2022lesicin}, judgment outcome prediction \cite{malik2021ildc}, judgment summarization \cite{bhattacharya2021}  provide AI building blocks. 
\cite{https://doi.org/10.48550/arxiv.2209.06049} created InLegalBERT and InCaseLawBERT, which are further pre-trained versions of LegalBERT \cite{chalkidis-etal-2020-legal} and CaseLawBERT \cite{zheng2021does} respectively on Indian legal text.

\section{Legal Named Entity Recognition Corpus}
\label{sec:NER_dataset}
\subsection{Legal Named Entities}
A typical Indian court judgment can be split into two parts viz., preamble and judgment. The preamble of a judgment contains formatted metadata like names of parties, judges, lawyers, date, court etc. The text following the preamble till the end of the judgment is called "judgment". An example showing the preamble and judgment of a court judgment along with entities is shown in Figure \ref{fig:NER_example}. The preamble typically ends with keywords like JUDGMENT or ORDER etc. In case these keywords are not found, we treat the first occurrence of 2 consecutive sentences with a verb as the start of the judgment part. This is because the preamble typically contains formatted metadata and not grammatically complete sentences.  

After discussion with legal experts about the useful information to be extracted from court judgments, we came up with a list of legal named entities which are described in Table \ref{ner_definitions}.
Some entities are extracted from the preamble, and some from the judgment text. Some entities are extracted from both the preamble and judgment, and their definitions may change depending on where they are extracted from. 

\begin{table*}[ht]
\centering
\begin{tabular}{ | m{7em} | m{1.5cm}| m{10.3cm} | }
    \hline
        \textbf{Named Entity} & \textbf{Extract From} & \textbf{Description} \\ \hline
        COURT & Preamble, Judgment & Name of the court which has delivered the current judgement if extracted from the preamble. Name of any court mentioned if extracted from  judgment sentences. \\ \hline
        PETITIONER & Preamble, Judgment & Name of the petitioners/appellants/revisionist  from current case \\ \hline
        RESPONDENT & Preamble, Judgment & Name of the respondents/defendants/opposition from current case \\ \hline
        JUDGE & Preamble, Judgment & Name of the judges from the current case  if extracted from the preamble. Name of the judges of the current as well as previous cases if extracted from judgment sentences.\\ \hline
        LAWYER & Preamble & Name of the lawyers from both the parties \\ \hline
        DATE & Judgment & Any date mentioned in the judgment \\ \hline
        ORG & Judgment & Name of organizations mentioned in text apart from the court. \\ \hline
        GPE & Judgment & Geopolitical locations which include names of states, cities, villages \\ \hline
        STATUTE & Judgment & Name of the act or law mentioned in the judgement \\ \hline
        PROVISION & Judgment & Sections, sub-sections, articles, orders, rules under a statute \\ \hline
        PRECEDENT & Judgment & All the past court cases referred to in the judgement as precedent. Precedent consists of party names + citation(optional) or case number (optional) \\ \hline
        CASE\_NUMBER & Judgment & All the other case numbers mentioned in the judgment (apart from precedent) where party names and citation is not provided \\ \hline
        WITNESS & Judgment & Name of witnesses in current judgment \\ \hline
        OTHER\_PERSON & Judgment & Name of all the persons that are not included in petitioner, respondent, judge and witness \\ \hline
\end{tabular}
\caption{\label{ner_definitions}
Legal Named Entities Definitions
}
\end{table*}

Flat entities were considered for annotation i.e., "Bank of China" should be considered as an ORG entity and "China" should not be marked as GPE inside this entity.
The detailed definitions with correctly and incorrectly marked examples can be found here\footnote{\url{https://storage.googleapis.com/indianlegalbert/OPEN_SOURCED_FILES/NER/NER_Definitions.pdf}}.

\subsection{Representative Sample of Indian High Court \& Supreme Court judgments}
Selecting a representative sample of court judgments text is vital to cover varieties of styles of writing judgments. Most cited judgements are likely to be more important for applying the NER model. But just taking the most cited judgments from a given court would produce bias in certain types of cases. Hence it is necessary to control case types as well.  We created the following 8 types of cases (Tax, Criminal, Civil, Motor Vehicles, Land \& Property, Industrial \& Labour, Constitution and Financial) which cover most of the cases in Indian courts. Classification of each judgment into one of these 8 types is a complex task. We have used a naive approach to use keywords based on act names for assigning a judgment to a case type. E.g., If the judgment mentions the "income tax act" then most probably it belongs to the "Tax" category. We use IndianKanoon search engine\footnote{\url{https://indiankanoon.org/}} to get the most cited court judgments matching the key act names. The key act names for each of the case types are given in Table \ref{table:act_names}.

\begin{table}
\centering
\begin{tabular}{ | m{5em} | m{5cm} |}
    \hline
        \textbf{Case Type} & \textbf{Key Act keywords} \\ \hline
        Tax & tax act, excise act, customs act, goods and services act etc. \\ \hline
        Criminal & IPC, penal code, criminal procedure etc. \\ \hline
        Civil & civil procedure, family courts, marriage act, wakf act etc. \\ \hline
        Motor Vehicles & motor vehicles act, mv act, imv act etc. \\ \hline
        Land \& Property & land acquisition act, succession act, rent control act etc. \\ \hline
        Industrial \& Labour & companies act, industrial disputes act, compensation act etc. \\ \hline
        Constitution & constitution \\ \hline
        Financial & negotiable instruments act, sarfaesi act, foreign exchange regulation act etc. \\ \hline
    \end{tabular}
\caption{\label{table:act_names}
Key Act Names for Each Case Type
}
\end{table}

One IndianKanoon search query was created for each of the 8 case types and 29 courts (supreme court, 23 high courts, three tribunals and 2 district courts). The Topmost cited results from each query were combined and de-duplicated to produce the final corpus of judgments. We consider judgments in the English language only. Judgments obtained by this method from 1950 to 2017 were used for training data annotations, and judgments from 2018 to March 2022 were used for the test and dev data annotations. The representative sample dataset of 11970 judgments, along with search queries, the full text of the judgments and descriptive statistics, are published in our git repository\footnote{\url{https://github.com/Legal-NLP-EkStep/legal_NER/tree/main/representative_judgments_sample}}. We believe these representative judgments can be used for other future studies as well.

\subsection{Data Annotation Process}
The annotations for judgment text were done at a sentence level, i.e. separate individual judgment sentences were presented for annotation without the document-level context. However, annotators had the freedom to access the entire judgment text by clicking on the Indiankanoon URL shown below the text in case they needed more context. Complete preambles were presented for annotation. 

\subsubsection{Selecting Raw Text to Annotate}
\label{subsec:raw_text_selection}
Legal named entities in a judgment text tend to be sparse, i.e., many of the sentences in a court judgment may not have any legal named entities. Hence is essential to identify entity-rich sentences for annotation rather than taking a random sample. We used the spacy pre-trained model (en\_core\_web\_trf) \cite{Montani2022-bt} with custom rules to predict the legal named entities. Custom rules were used to map the Spacy-defined named entities to the legal named entities defined in this paper. E.g., An entity predicted by spacy as PERSON with the keyword "petitioner" nearby was marked as PETITIONER etc.
We passed the representative sample judgment texts through this Spacy model with custom rules to get predicted noisy legal entities. Using these predicted legal entities, we selected the sentences that are entity-rich and that reduce the class imbalance across different entity types. We also added sentences without any predicted entities. Very short sentences and sentences with non-English characters were discarded. Preambles, where party names are written side by side on the same line, were also discarded. 

\subsubsection{Pre-annotations}
The data annotation was done in 4 cycles. The preambles and sentences were pre-annotated in each cycle to reduce annotation effort.

For the first annotation cycle, the predicted legal entities obtained during the raw text selection process, as mentioned in \ref{subsec:raw_text_selection}, were reviewed and corrected. At the end of cycle 1, a machine learning model using Roberta+ transition-based parser architecture (explained in detail in \S \ref{sec:ner_baseline_model}) was trained using the labelled data obtained in cycle 1. This machine learning model was used to pre-annotate the cycle 2 data. Similarly, the machine learning model trained using cycle 1 and 2 data was used to pre-annotate cycle 3 data and so on. 

\subsubsection{Manual Reviews \& Corrections}
In each cycle, all of the pre-annotated preambles and sentences were carefully reviewed and corrected by humans. Roughly the same amount of preamble and sentences were annotated in each cycle. The team of 4 legal experts and 4 data scientists at OpenNyAI did the data annotation. Legal experts were law students from various law universities across India.
We did not do duplicate annotations to maximize the number of annotated data. 
We used the Prodigy tool\footnote{\url{https://prodi.gy/}} for the annotations.

The corrected data obtained from the four annotation cycles was split into the train, dev and test datasets as per the time ranges mentioned in Table \ref{table:train_test}. We tried to keep the dev data distribution similar to the test data distribution. Test and dev data was carefully cross-reviewed twice to ensure data quality. 
The total count of entities, total number of preambles and judgment sentences in train, dev and test data are shown in Table \ref{table:train_test}.
\begin{table}[ht]
    \begin{center}
    \begin{tabular}{| m{5em} | m{1.2cm} | m{1.2cm} | m{1.2cm}|}
    \hline
        \textbf{} & \textbf{Train} & \textbf{Dev} & \textbf{Test} \\ \hline
        Time Range & 1950 to 2017 & 2018 to 2022 & 2018 to 2022 \\ \hline
        Preambles & 1560 & 125 & 441 \\ \hline
        Judgment sentences & 9435 & 949 & 4060 \\ \hline
        Entities & 29964 & 3216 & 13365 \\ \hline
    \end{tabular}
    \end{center}
\caption{\label{table:train_test}
Train \& Test data counts
}
\end{table}
The counts of each legal named entity in training data are shown in Table \ref{table:ner_cnt}.
\begin{table}[ht]
    \centering
    \begin{tabular}{ | m{7em} | m{1.4cm} | m{1.4cm} |}
    \hline
        \textbf{Entity} & \textbf{Judgment Count} & \textbf{Preamble Count} \\ \hline
        COURT & 1293 & 1074 \\ \hline
        PETITIONER & 464 & 2604 \\ \hline
        RESPONDENT & 324 & 3538 \\ \hline
        JUDGE & 567 & 1758 \\ \hline
        LAWYER & NA & 3505 \\ \hline
        DATE & 1885 & NA \\ \hline
        ORG & 1441 & NA \\ \hline
        GPE & 1398 & NA \\ \hline
        STATUTE & 1804 & NA \\ \hline
        PROVISION & 2384 & NA \\ \hline
        PRECEDENT & 1351 & NA \\ \hline
        CASE\_NUMBER & 1040 & NA \\ \hline
        WITNESS & 881 & NA \\ \hline
        OTHER\_PERSON & 2653 & NA \\ \hline \hline
        \textbf{Total} & 17485 & 12479 \\ \hline
    \end{tabular}
\caption{\label{table:ner_cnt}
Counts of Legal Entities for Training data in Preamble \& Judgment
}
\end{table}
\section{NER Baseline Model}
\label{sec:ner_baseline_model}
\begin{figure}[ht]
\begin{center}
\includegraphics[scale=0.6]{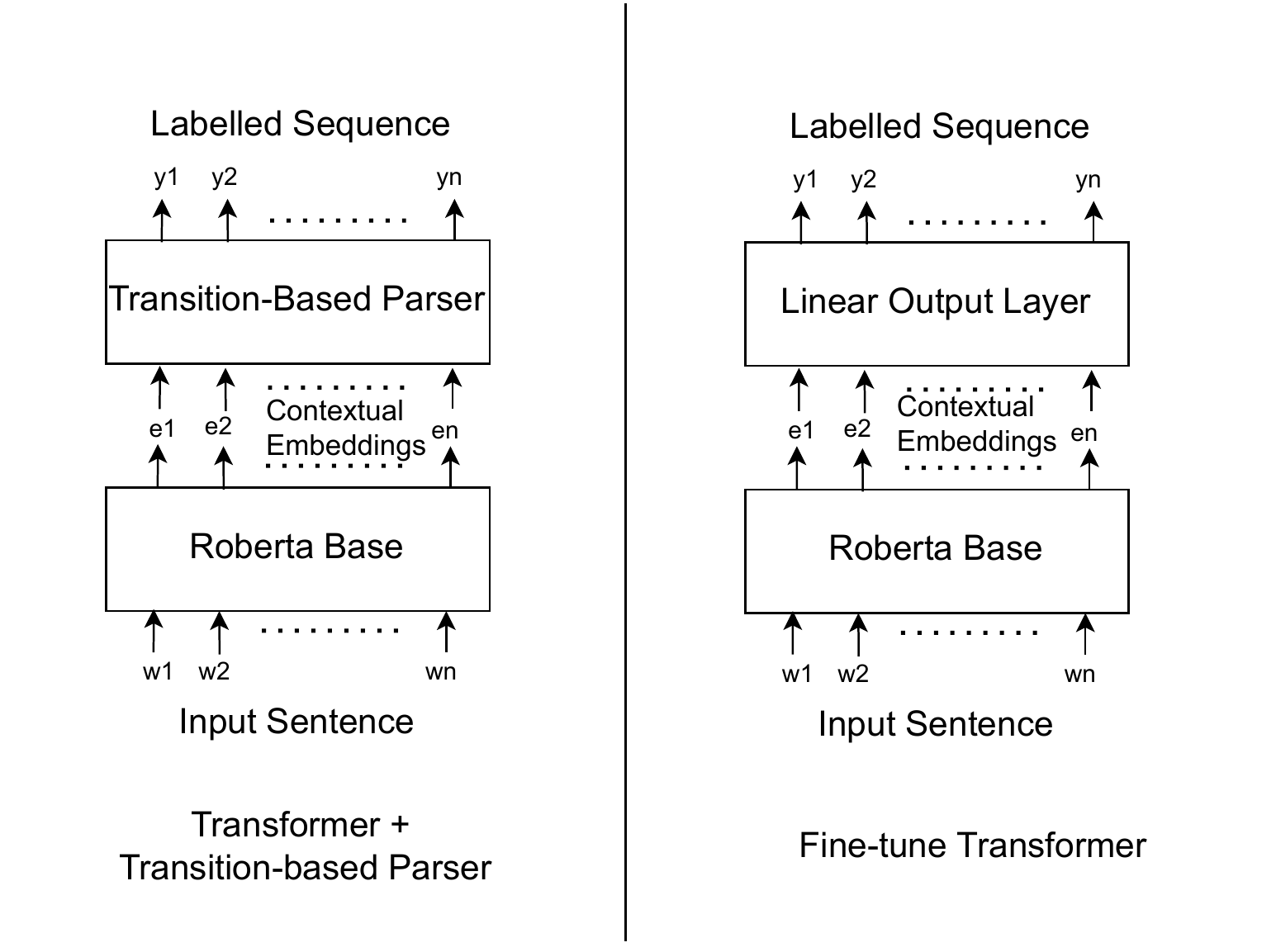} 
\caption{NER Architectures}
\label{fig:NER_architecture}
\end{center}
\end{figure}

The end goal behind this work is to enable the development of other legal AI applications that consume automatically detected legal named entities from judgment texts. Towards this goal, we experimented with some famous NER model architectures. A single model was trained to predict entities from both judgment sentences and the preamble. As transformer-based architectures have shown a lot of success in NER tasks \cite{li2020survey}, we mainly experimented with them. We compared the performance of 2 NER architecture types when trained on our legal NER dataset. The first architecture type uses a transition-based dependency parser \cite{honnibal2015improved} on top of the transformer model. The second architecture type uses a fine-tuning based approach which adds a single linear layer to the transformer model and fine-tunes the entire architecture on the NER task. Figure \ref{fig:NER_architecture} shows the 2 NER architecture types.

We experimented with multiple transformer models for each of the architecture types. For transition-based parser architecture we experimented with the Roberta-base model \cite{liu2019roberta}, InLegalBERT \cite{https://doi.org/10.48550/arxiv.2209.06049} using Spacy library. For the fine-tuning approach we experimented with Roberta-base, InLegalBERT, legalBERT \cite{chalkidis-etal-2020-legal} using TNER library \cite{ushio-camacho-collados-2021-ner}. 

The models are evaluated by using recall, precision and strict F1 scores on combined preamble and judgment sentences. The named entity is considered correct when both boundary and entity class are predicted correctly.
Table \ref{table:ner_experiments} shows the performance of these experiments on the test data. 
\begin{table}
\centering
\begin{tabular}{ | m{5.2em} | m{1.2cm} | m{0.65cm}| m{0.65cm}| m{0.7cm}|}
        \hline
        \textbf{Architecture Type} & \textbf{Trans. Model} & \textbf{P} & \textbf{R} & \textbf{F1} \\ \hline
        \multirow{2}{6em}{Transformer + Transition Based Parser} & \textbf{Roberta-base} & \textbf{92.0} & \textbf{90.2} & \textbf{91.1} \\ \cline{2-5}
        & InLegal BERT & 87.3 & 85.8 & 86.5 \\ \hline
        \multirow{3}{6em}{Fine Tune Transformer} & Roberta-base & 77.6 & 80.0 & 78.8 \\ \cline{2-5}
        & InLegal BERT & 77.7 & 84.6 & 81.0 \\\cline{2-5}
        & Legal BERT & 75.4 & 79.5 & 77.5 \\
        \hline
    \end{tabular}
\caption{\label{table:ner_experiments}
Model Performance on test data
}
\end{table}

Performance of the best performing model (Roberta+ transition-based parser) on each of the entity classes on test data along with average character length is shown in Table \ref{table:entity_wise_results}. It also shows the Type match F1 score which was proposed in \cite{segura-bedmar-etal-2013-semeval}. Under the Type match evaluation scheme, some overlap between the tagged entity and the gold entity is required along with entity type match. 
In strict f1 calculation, the entities with correct entity type match and partial span overlap are considered incorrect. But in the Type match evaluation, such entities are considered the correct entity. Hence the Type match F1 score gives an indication of how much overlap exists between ground truth and prediction considering partial matches.
\begin{table}
\centering
\begin{tabular}{ | m{7em} | m{0.8cm} |m{0.6cm} |m{0.65cm} | m{0.75cm} |}
        \hline
        \textbf{Entity} & \textbf{Count} & \textbf{Avg. Len.} & \textbf{F1} & \textbf{Type match F1}\\ \hline
        COURT & 1231 & 25 & 95.4 & 97.2 \\ \hline
        PETITIONER & 835 & 20 & 89.8 & 92.6 \\ \hline
        RESPONDENT & 1125 & 34 & 83.0 & 91.8 \\ \hline
        JUDGE & 580 & 15 & 95.4 & 96.5 \\ \hline
        LAWYER & 1813 & 16 & 94.1 & 95.5 \\ \hline
        DATE & 1111 & 11 & 91.9 & 98.7 \\ \hline
        ORG & 920 & 18 & 86.4 & 90.2 \\ \hline
        GPE & 711 & 8 & 85.7 & 90.9 \\ \hline
        STATUTE & 971 & 17 & 96.0 & 97.6 \\ \hline
        PROVISION & 1220 & 14 & 95.7 & 98.6 \\ \hline
        PRECEDENT & 634 & 62 & 80.1 & 96.2 \\ \hline
        CASE\_NUMBER & 683 & 23 & 89.1 & 92.4 \\ \hline
        WITNESS & 446 & 12 & 89.7 & 89.7 \\ \hline
        OTHER\_PERSON & 1085 & 12 & 93.8 & 95.2 \\ \hline        
        \textbf{Overall} & \textbf{13365} & \textbf{20} & \textbf{91.1} & \textbf{94.9} \\ \hline
        
    \end{tabular}
\caption{\label{table:entity_wise_results}
Entity-wise performance of Roberta + Transition-based Parser model on test data
}
\end{table}

The trained Roberta+Transition-based Parser is made available as a Spacy pipeline in our git repository and hugging face model repository\footnote{\url{https://huggingface.co/opennyaiorg/en_legal_ner_trf}}. 

\subsection{Training Procedure}
\label{appendix:training_procedure}
Early stopping using dev data was used during training to select the best epoch for all the experiments.
The details about the training procedure for Roberta + Transition-based parser using Spacy are available in the GitHub repository.  NVIDIA Tesla V100 GPU was used to train the model and the training time was 12 hours. The key parameters used are mentioned in Table \ref{table:training_procedure}.

\begin{table}
\centering
\begin{tabular}{ | m{7.5em} | m{4cm} |}
    \hline
        \textbf{Parameter} & \textbf{Value} \\ \hline
        Transformer & Roberta-base \\ \hline
        Optimizer & Adam with beta1 = 0.9, beta2 = 0.999, L2 = 0.01, initial learning rate = 0.00005  \\ \hline
        max training steps & 40000 \\ \hline
        training batch size & 256 \\ \hline
    \end{tabular}
\caption{\label{table:training_procedure}
Key training procedure parameters
}
\end{table}

\subsection{Results Discussion}
Adding a transition-based parser to the transformer architecture significantly improves the model's accuracy for this NER task, as seen in Table \ref{table:ner_experiments}.  

As seen from Table \ref{table:entity_wise_results}, the Roberta-base + transition-based parser NER model can extract shorter entities like WITNESS, PROVISION, STATUTE, LAWYER, COURT and JUDGE with excellent performance.

PRECEDENT has degraded performance as compared to other entities because the precedent names are usually very long (average entity length is 62 characters) and missing out on even a few characters makes the entire entity to be marked as incorrect. Because of this reason, there is a significant difference between strict F1 and Type match F1 for PRECEDENT. Manual inspection of errors in PRECEDENT prediction reveals that many a time, the prefixes of party names like "Mr.", "M/S", etc.  are missed in the prediction while gold entities have them. E.g., the gold entity type is PRECEDENT with the text "Mr Amit Kumar Vs State of Maharashtra," and the predicted entity type is PRECEDENT with the text "Amit Kumar Vs State of Maharashtra". In strict F1 evaluation, this example is considered incorrect, while in Type match evaluation, this example is considered correct. 
One possible reason for the model not to include the prefixes in the PRECEDENT prediction could be that prefixes are not considered as a part of other entities like PETITIONER, RESPONDENT and ORG. So possibly, the model has also learned to omit the prefixes in PRECEDENT. 

The average character length of RESPONDENT entities is considerably higher than that of PETITIONER entities.
This difference is because, often, the respondents are posts or authorities rather than a person. E.g. "The Chief Engineer, Water Resource Organization, Chepauk, Chennai-5". In such cases, gold data marks the authority or post names along with the address as the corresponding entity, making them longer. The difference between strict F1 and Type match F1 for RESPONDENT shows that the model is missing in predicting a few characters in such long entities.

The overall accuracy of this model makes it very useful in practical legal AI applications.
\section{Post-Processing of Named Entities}
 Since the annotators were asked to annotate individual sentences without document-level context, any trained NER model on this data will also focus only on the sentence-level information.
 While inferring the NER model on a complete judgment text, it is important to perform post-processing of extracted entities to capture document-level context. In particular, we create rules to perform the following tasks
\begin{itemize}
\item Reconciliation of the entities extracted from individual sentences of a judgment
\item Coreference resolution of precedents
\item Coreference resolution of statutes
\item Assign statute to every provision
\end{itemize}

 \subsection{Reconciliation of the Extracted Entities}
 The same entity text can be tagged with different legal entity classes in separate sentences of the same judgment. 
 E.g., In the preamble of a judgment, it is written that "Amit Kumar" is a petitioner. In the same judgement text, the judge later writes, "Four unidentified persons attacked Amit Kumar". NER model would mark "Amit Kumar" in the second mention as OTHER\_PERSON because there is no information about Amit Kumar being a petitioner in this sentence. Marking this person's name as PETITIONER is more valuable than marking it as an OTHER\_PERSON. 

As part of entity reconciliation, entities predicted as OTHER\_PERSON or ORG are matched with all the PETITIONER, RESPONDENT, JUDGE, LAWYER and WITNESS entities. If an exact match is found, then the entity type is overwritten with the matching entity type. In the previous example, all the extracted entities that match "Amit Kumar" would be overwritten with entity type PETITIONER. 

 \subsection{Coreference Resolution of Precedents}
The names of precedent cases are usually very long. Hence judges typically mention the complete name of a precedent case for the first mention and later use the name of the first party as a reference. E.g., "The constitution bench of this court in Gurbaksh Singh Sibbia and others Vs State of Punjab (1980) 2 SCC 565 dealt with the scope and ambit of anticipatory bail". Then, later on, the judge uses a reference to this case, like "The learned counsel for the petitioner placed reliance on Sibbia's case (supra)."
The NER model identifies "Sibbia" as OTHER\_PERSON in the second sentence. But here, "Sibbia" is a reference to the earlier extracted PRECEDENT entity "Gurbaksh Singh Sibbia and others Vs State of Punjab (1980) 2 SCC 565".

We first cluster all the extracted precedent entities within a judgment by matching the party names and citations. A precedent cluster contains all the precedent entities with matching party names or citations. Then we identify potential precedent referents as ORG or OTHER\_PERSON entity followed by keywords "supra" or "'s case". We then search such referent entities in the extracted precedents' party names and find the closest matching preceding precedent. If the match is found, then we change the referent entity type to PRECEDENT. Referent entities are also added to the precedent cluster where the closest matching precedent belongs. Once all the matching precedent referents are assigned to precedent clusters, the longest entity in each cluster is marked as the cluster head. 
So in the example before, the entity type for "Sibbia" in the second sentence would be changed from OTHER\_PERSON to PRECEDENT, and a precedent cluster would be created with the head as "Gurbaksh Singh Sibbia and others Vs State of Punjab (1980) 2 SCC 565" and the member as "Sibbia".

The information about precedents coreference can be accessed through output Spacy doc object property \emph{doc.user\_data['precedent\_clusters']}.

 \subsection{Coreference Resolution of Statutes}
Statute names can be long and are frequently mentioned in judgment text. Hence judges typically write the complete statute name at the beginning of the judgment and specify the referent for this statute for the remaining judgement. E.g., "The complaint was filed under the Companies Act, 1956 (for brevity, 'the Act') ...". Later on in the same judgment, the judge writes ", Section 5 of the Act defines ...".  We write rules to identify such statute referents by searching for a STATUTE entity followed by keywords in parenthesis. Such statute referents are added to the statute cluster with its head as the complete statute name. All entities in a statute cluster refer to the same statute, which is the head of the cluster. Extracted statutes are also looked up against a list of famous acronyms (IPC, CrPC etc.), and if a match is found, then the corresponding full form is added to the statutes cluster. The information about statute coreference can be accessed through output Spacy doc object property \emph{doc.user\_data['statute\_clusters']}.
 
 \subsection{Assign Statute to Every Provision}
 Every extracted provision should be associated with an extracted statute. Sometimes a provision and its corresponding statute are explicitly mentioned in the same sentence. E.g., "Section 420 of Indian Penal Code says ...". Sometimes, the provision-statute mapping is implicit where only the provision is mentioned, and the corresponding statute is understood from the context. E.g., "The section 420 says ...". 
 
In case of explicit mentions, we assign the statute to the immediately preceding provision in the same sentence. All the remaining provisions are considered implicit provisions. We first search for all the implicit provisions if a unique explicit mapping exists in another sentence. E.g., if the judge writes an explicit mention like "Section 420 of Indian Penal Code" and there is no other explicit mention of Section 420 for any other statute in the entire judgment text, then all the implicit mentions of Section 420 are mapped to "Indian Penal Code". Suppose no explicit mention for a provision is found, or multiple explicit mentions are found for a provision. In that case, the statute extracted from the closest preceding sentence is assigned.

The assignment of the statute to provisions can be accessed via output Spacy doc object property \emph{doc.user\_data['provision\_statute\_pairs']}
\section{Conclusion \& Future Directions}
In this paper, we proposed a new corpus of legal named  entities using 14 legal entity types. We also proposed baseline models trained using this corpus along with post-processing of the extracted entities to capture document-level information. We have also released a representative sample of Indian court judgments which could be used in further studies. We believe this corpus will lay the foundation for further NLP tasks like relationship extraction, knowledge graph population etc. using Indian court judgments.

\section*{Acknowledgements}
This work is part of the OpenNyAI mission, which is funded by EkStep and Agami. We thank all the law experts, student volunteers for contributing to data annotation.


\appendix

\bibliography{anthology,custom}

\begin{thebibliography}{37}
\expandafter\ifx\csname natexlab\endcsname\relax\def\natexlab#1{#1}\fi

\bibitem[{Angelidis et~al.(2018)Angelidis, Chalkidis, and
  Koubarakis}]{angelidis2018named}
Iosif Angelidis, Ilias Chalkidis, and Manolis Koubarakis. 2018.
\newblock Named entity recognition, linking and generation for greek
  legislation.
\newblock In \emph{JURIX}, pages 1--10.

\bibitem[{Barriere and Fouret(2019)}]{barriere2019may}
Valentin Barriere and Amaury Fouret. 2019.
\newblock May i check again?—a simple but efficient way to generate and use
  contextual dictionaries for named entity recognition. application to french
  legal texts.
\newblock In \emph{Proceedings of the 22nd Nordic Conference on Computational
  Linguistics}, pages 327--332.

\bibitem[{Benikova et~al.(2015)Benikova, Muhie, Prabhakaran, and
  Biemann}]{benikova2015c}
Darina Benikova, Seid Muhie, Yimam Prabhakaran, and Santhanam~Chris Biemann.
  2015.
\newblock C.: Germaner: Free open german named entity recognition tool.
\newblock In \emph{In: Proc. GSCL-2015}. Citeseer.

\bibitem[{Bikel et~al.(1999)Bikel, Schwartz, and
  Weischedel}]{bikel1999algorithm}
Daniel~M Bikel, Richard Schwartz, and Ralph~M Weischedel. 1999.
\newblock An algorithm that learns what's in a name.
\newblock \emph{Machine learning}, 34(1):211--231.

\bibitem[{Borthwick et~al.(1998)Borthwick, Sterling, Agichtein, and
  Grishman}]{borthwick1998nyu}
Andrew Borthwick, John Sterling, Eugene Agichtein, and Ralph Grishman. 1998.
\newblock Nyu: Description of the mene named entity system as used in muc-7.
\newblock In \emph{Seventh Message Understanding Conference (MUC-7):
  Proceedings of a Conference Held in Fairfax, Virginia, April 29-May 1, 1998}.

\bibitem[{Cardellino et~al.(2017)Cardellino, Teruel, Alemany, and
  Villata}]{cardellino2017low}
Cristian Cardellino, Milagro Teruel, Laura~Alonso Alemany, and Serena Villata.
  2017.
\newblock A low-cost, high-coverage legal named entity recognizer, classifier
  and linker.
\newblock In \emph{Proceedings of the 16th edition of the International
  Conference on Articial Intelligence and Law}, pages 9--18.

\bibitem[{Chalkidis et~al.(2020)Chalkidis, Fergadiotis, Malakasiotis, Aletras,
  and Androutsopoulos}]{chalkidis-etal-2020-legal}
Ilias Chalkidis, Manos Fergadiotis, Prodromos Malakasiotis, Nikolaos Aletras,
  and Ion Androutsopoulos. 2020.
\newblock \href {https://doi.org/10.18653/v1/2020.findings-emnlp.261}
  {{LEGAL}-{BERT}: The muppets straight out of law school}.
\newblock In \emph{Findings of the Association for Computational Linguistics:
  EMNLP 2020}, pages 2898--2904, Online. Association for Computational
  Linguistics.

\bibitem[{Chalkidis et~al.(2021)Chalkidis, Fergadiotis, Malakasiotis, and
  Androutsopoulos}]{chalkidis2021neural}
Ilias Chalkidis, Manos Fergadiotis, Prodromos Malakasiotis, and Ion
  Androutsopoulos. 2021.
\newblock Neural contract element extraction revisited: Letters from sesame
  street.
\newblock \emph{arXiv preprint arXiv:2101.04355}.

\bibitem[{Daiber et~al.(2013)Daiber, Jakob, Hokamp, and
  Mendes}]{daiber2013improving}
Joachim Daiber, Max Jakob, Chris Hokamp, and Pablo~N Mendes. 2013.
\newblock Improving efficiency and accuracy in multilingual entity extraction.
\newblock In \emph{Proceedings of the 9th international conference on semantic
  systems}, pages 121--124.

\bibitem[{Dozier et~al.(2010)Dozier, Kondadadi, Light, Vachher, Veeramachaneni,
  and Wudali}]{dozier2010named}
Christopher Dozier, Ravikumar Kondadadi, Marc Light, Arun Vachher, Sriharsha
  Veeramachaneni, and Ramdev Wudali. 2010.
\newblock Named entity recognition and resolution in legal text.
\newblock In \emph{Semantic Processing of Legal Texts}, pages 27--43. Springer.

\bibitem[{Glaser et~al.(2018)Glaser, Waltl, and Matthes}]{glaser2018named}
Ingo Glaser, Bernhard Waltl, and Florian Matthes. 2018.
\newblock Named entity recognition, extraction, and linking in german legal
  contracts.
\newblock In \emph{IRIS: Internationales Rechtsinformatik Symposium}, pages
  325--334.

\bibitem[{Honnibal and Johnson(2015)}]{honnibal2015improved}
Matthew Honnibal and Mark Johnson. 2015.
\newblock An improved non-monotonic transition system for dependency parsing.
\newblock In \emph{Proceedings of the 2015 conference on empirical methods in
  natural language processing}, pages 1373--1378.

\bibitem[{Joshi et~al.(2020)Joshi, Chen, Liu, Weld, Zettlemoyer, and
  Levy}]{joshi2020spanbert}
Mandar Joshi, Danqi Chen, Yinhan Liu, Daniel~S Weld, Luke Zettlemoyer, and Omer
  Levy. 2020.
\newblock Spanbert: Improving pre-training by representing and predicting
  spans.
\newblock \emph{Transactions of the Association for Computational Linguistics},
  8:64--77.

\bibitem[{Kalamkar et~al.(2022)Kalamkar, Tiwari, Agarwal, Karn, Gupta,
  Raghavan, and Modi}]{kalamkar-EtAl:2022:LREC}
Prathamesh Kalamkar, Aman Tiwari, Astha Agarwal, Saurabh Karn, Smita Gupta,
  Vivek Raghavan, and Ashutosh Modi. 2022.
\newblock \href {https://aclanthology.org/2022.lrec-1.470} {Corpus for
  automatic structuring of legal documents}.
\newblock In \emph{Proceedings of the Language Resources and Evaluation
  Conference}, pages 4420--4429, Marseille, France. European Language Resources
  Association.

\bibitem[{Lample et~al.(2016)Lample, Ballesteros, Subramanian, Kawakami, and
  Dyer}]{lample2016neural}
Guillaume Lample, Miguel Ballesteros, Sandeep Subramanian, Kazuya Kawakami, and
  Chris Dyer. 2016.
\newblock Neural architectures for named entity recognition.
\newblock In \emph{Proceedings of the 2016 Conference of the North American
  Chapter of the Association for Computational Linguistics: Human Language
  Technologies}, pages 260--270.

\bibitem[{Leitner et~al.(2020)Leitner, Rehm, and
  Schneider}]{leitner2020dataset}
Elena Leitner, Georg Rehm, and Julian~Moreno Schneider. 2020.
\newblock A dataset of german legal documents for named entity recognition.
\newblock In \emph{Proceedings of the 12th Language Resources and Evaluation
  Conference}, pages 4478--4485.

\bibitem[{Li et~al.(2020)Li, Sun, Han, and Li}]{li2020survey}
Jing Li, Aixin Sun, Jianglei Han, and Chenliang Li. 2020.
\newblock A survey on deep learning for named entity recognition.
\newblock \emph{IEEE Transactions on Knowledge and Data Engineering},
  34(1):50--70.

\bibitem[{Liu et~al.(2019)Liu, Ott, Goyal, Du, Joshi, Chen, Levy, Lewis,
  Zettlemoyer, and Stoyanov}]{liu2019roberta}
Yinhan Liu, Myle Ott, Naman Goyal, Jingfei Du, Mandar Joshi, Danqi Chen, Omer
  Levy, Mike Lewis, Luke Zettlemoyer, and Veselin Stoyanov. 2019.
\newblock Roberta: A robustly optimized bert pretraining approach.
\newblock \emph{arXiv preprint arXiv:1907.11692}.

\bibitem[{Luz~de Araujo et~al.(2018)Luz~de Araujo, Campos, de~Oliveira,
  Stauffer, Couto, and Bermejo}]{luz2018lener}
Pedro~Henrique Luz~de Araujo, Te{\'o}filo E~de Campos, Renato~RR de~Oliveira,
  Matheus Stauffer, Samuel Couto, and Paulo Bermejo. 2018.
\newblock Lener-br: a dataset for named entity recognition in brazilian legal
  text.
\newblock In \emph{International Conference on Computational Processing of the
  Portuguese Language}, pages 313--323. Springer.

\bibitem[{Malik et~al.(2021)Malik, Sanjay, Nigam, Ghosh, Guha, Bhattacharya,
  and Modi}]{malik2021ildc}
Vijit Malik, Rishabh Sanjay, Shubham~Kumar Nigam, Kripabandhu Ghosh,
  Shouvik~Kumar Guha, Arnab Bhattacharya, and Ashutosh Modi. 2021.
\newblock Ildc for cjpe: Indian legal documents corpus for court judgment
  prediction and explanation.
\newblock In \emph{Proceedings of the 59th Annual Meeting of the Association
  for Computational Linguistics and the 11th International Joint Conference on
  Natural Language Processing (Volume 1: Long Papers)}, pages 4046--4062.

\bibitem[{McCallum and Li(2003)}]{mccallum2003early}
Andrew McCallum and Wei Li. 2003.
\newblock Early results for named entity recognition with conditional random
  fields, feature induction and web-enhanced lexicons.
\newblock In \emph{Proceedings of the seventh conference on Natural language
  learning at HLT-NAACL 2003-Volume 4}, pages 188--191.

\bibitem[{Mendes et~al.(2011)Mendes, Jakob, Garc{\'\i}a-Silva, and
  Bizer}]{mendes2011dbpedia}
Pablo~N Mendes, Max Jakob, Andr{\'e}s Garc{\'\i}a-Silva, and Christian Bizer.
  2011.
\newblock Dbpedia spotlight: shedding light on the web of documents.
\newblock In \emph{Proceedings of the 7th international conference on semantic
  systems}, pages 1--8.

\bibitem[{Montani et~al.(2022)Montani, Honnibal, Honnibal, Van~Landeghem, Boyd,
  Peters, McCann, Samsonov, Geovedi, O'Regan, Altinok, Orosz, Kristiansen,
  {Roman}, Bot, Miranda, Fiedler, de~Kok, Howard, {Edward}, Phatthiyaphaibun,
  Tamura, Bozek, {murat}, Amery, Daniels, B{\"o}ing, Tippa, and
  Baumgartner}]{Montani2022-bt}
Ines Montani, Matthew Honnibal, Matthew Honnibal, Sofie Van~Landeghem, Adriane
  Boyd, Henning Peters, Paul~O'leary McCann, Maxim Samsonov, Jim Geovedi, Jim
  O'Regan, Duygu Altinok, Gy{\"o}rgy Orosz, S{\o}ren~Lind Kristiansen, {Roman},
  Explosion Bot, Lj~Miranda, Leander Fiedler, Dani{\"e}l de~Kok, Gr{\'e}gory
  Howard, {Edward}, Wannaphong Phatthiyaphaibun, Yohei Tamura, Sam Bozek,
  {murat}, Mark Amery, Ryn Daniels, Bj{\"o}rn B{\"o}ing, Pradeep~Kumar Tippa,
  and Peter Baumgartner. 2022.
\newblock explosion/spacy: v3.2.4: Workaround for {Click/Typer} issues.

\bibitem[{Ouchi et~al.(2020)Ouchi, Suzuki, Kobayashi, Yokoi, Kuribayashi,
  Konno, and Inui}]{ouchi2020instance}
Hiroki Ouchi, Jun Suzuki, Sosuke Kobayashi, Sho Yokoi, Tatsuki Kuribayashi,
  Ryuto Konno, and Kentaro Inui. 2020.
\newblock Instance-based learning of span representations: A case study through
  named entity recognition.
\newblock In \emph{Proceedings of the 58th Annual Meeting of the Association
  for Computational Linguistics}, pages 6452--6459.

\bibitem[{P{\u{a}}iș et~al.(2021)P{\u{a}}iș, Mitrofan, Gasan, Coneschi, and
  Ianov}]{puais2021named}
Vasile P{\u{a}}iș, Maria Mitrofan, Carol~Luca Gasan, Vlad Coneschi, and
  Alexandru Ianov. 2021.
\newblock Named entity recognition in the romanian legal domain.
\newblock In \emph{Proceedings of the Natural Legal Language Processing
  Workshop 2021}, pages 9--18.

\bibitem[{Paul et~al.(2022{\natexlab{a}})Paul, Goyal, and
  Ghosh}]{paul2022lesicin}
Shounak Paul, Pawan Goyal, and Saptarshi Ghosh. 2022{\natexlab{a}}.
\newblock Lesicin: A heterogeneous graph-based approach for automatic legal
  statute identification from indian legal documents.
\newblock In \emph{Proceedings of the AAAI Conference on Artificial
  Intelligence}, volume~36, pages 11139--11146.

\bibitem[{Paul et~al.(2022{\natexlab{b}})Paul, Mandal, Goyal, and
  Ghosh}]{https://doi.org/10.48550/arxiv.2209.06049}
Shounak Paul, Arpan Mandal, Pawan Goyal, and Saptarshi Ghosh.
  2022{\natexlab{b}}.
\newblock \href {https://doi.org/10.48550/ARXIV.2209.06049} {Pre-training
  transformers on indian legal text}.

\bibitem[{Schneider et~al.(2020)Schneider, Rehm, Montiel-Ponsoda, Doncel,
  Revenko, Karampatakis, Khvalchik, Sageder, Gracia, and
  Maganza}]{schneider2020orchestrating}
Julian~Moreno Schneider, Georg Rehm, Elena Montiel-Ponsoda,
  V{\'\i}ctor~Rodr{\'\i}guez Doncel, Artem Revenko, Sotirios Karampatakis,
  Maria Khvalchik, Christian Sageder, Jorge Gracia, and Filippo Maganza. 2020.
\newblock Orchestrating nlp services for the legal domain.
\newblock In \emph{Proceedings of the 12th Language Resources and Evaluation
  Conference}, pages 2332--2340.

\bibitem[{Segura-Bedmar et~al.(2013)Segura-Bedmar, Mart{\'\i}nez, and
  Herrero-Zazo}]{segura-bedmar-etal-2013-semeval}
Isabel Segura-Bedmar, Paloma Mart{\'\i}nez, and Mar{\'\i}a Herrero-Zazo. 2013.
\newblock \href {https://aclanthology.org/S13-2056} {{S}em{E}val-2013 task 9 :
  Extraction of drug-drug interactions from biomedical texts ({DDIE}xtraction
  2013)}.
\newblock In \emph{Second Joint Conference on Lexical and Computational
  Semantics (*{SEM}), Volume 2: Proceedings of the Seventh International
  Workshop on Semantic Evaluation ({S}em{E}val 2013)}, pages 341--350, Atlanta,
  Georgia, USA. Association for Computational Linguistics.

\bibitem[{Shukla et~al.(2022)Shukla, Bhattacharya, Poddar, Mukherjee, Ghosh,
  Goyal, and Ghosh}]{bhattacharya2021}
Abhay Shukla, Paheli Bhattacharya, Soham Poddar, Rajdeep Mukherjee, Kripabandhu
  Ghosh, Pawan Goyal, and Saptarshi Ghosh. 2022.
\newblock Legal case document summarization: Extractive and abstractive methods
  and their evaluation.
\newblock In \emph{The 2nd Conference of the Asia-Pacific Chapter of the
  Association for Computational Linguistics and the 12th International Joint
  Conference on Natural Language Processing}.

\bibitem[{Ushio and Camacho-Collados(2021)}]{ushio-camacho-collados-2021-ner}
Asahi Ushio and Jose Camacho-Collados. 2021.
\newblock \href {https://doi.org/10.18653/v1/2021.eacl-demos.7} {{T}-{NER}: An
  all-round python library for transformer-based named entity recognition}.
\newblock In \emph{Proceedings of the 16th Conference of the European Chapter
  of the Association for Computational Linguistics: System Demonstrations},
  pages 53--62, Online. Association for Computational Linguistics.

\bibitem[{Wang et~al.(2021)Wang, Gao, Zhu, Zhang, Liu, Li, and
  Tang}]{wang2021kepler}
Xiaozhi Wang, Tianyu Gao, Zhaocheng Zhu, Zhengyan Zhang, Zhiyuan Liu, Juanzi
  Li, and Jian Tang. 2021.
\newblock Kepler: A unified model for knowledge embedding and pre-trained
  language representation.
\newblock \emph{Transactions of the Association for Computational Linguistics},
  9:176--194.

\bibitem[{Yamada et~al.(2020)Yamada, Asai, Shindo, Takeda, and
  Matsumoto}]{yamada2020luke}
Ikuya Yamada, Akari Asai, Hiroyuki Shindo, Hideaki Takeda, and Yuji Matsumoto.
  2020.
\newblock Luke: Deep contextualized entity representations with entity-aware
  self-attention.
\newblock In \emph{Proceedings of the 2020 Conference on Empirical Methods in
  Natural Language Processing (EMNLP)}, pages 6442--6454.

\bibitem[{Ye et~al.(2022)Ye, Lin, Li, and Sun}]{ye2022packed}
Deming Ye, Yankai Lin, Peng Li, and Maosong Sun. 2022.
\newblock Packed levitated marker for entity and relation extraction.
\newblock In \emph{Proceedings of the 60th Annual Meeting of the Association
  for Computational Linguistics (Volume 1: Long Papers)}, pages 4904--4917.

\bibitem[{Yu et~al.(2020)Yu, Bohnet, and Poesio}]{yu2020named}
Juntao Yu, Bernd Bohnet, and Massimo Poesio. 2020.
\newblock Named entity recognition as dependency parsing.
\newblock In \emph{Proceedings of the 58th Annual Meeting of the Association
  for Computational Linguistics}, pages 6470--6476.

\bibitem[{Zheng et~al.(2021)Zheng, Guha, Anderson, Henderson, and
  Ho}]{zheng2021does}
Lucia Zheng, Neel Guha, Brandon~R Anderson, Peter Henderson, and Daniel~E Ho.
  2021.
\newblock When does pretraining help? assessing self-supervised learning for
  law and the casehold dataset of 53,000+ legal holdings.
\newblock In \emph{Proceedings of the Eighteenth International Conference on
  Artificial Intelligence and Law}, pages 159--168.

\bibitem[{Zhong et~al.(2020)Zhong, Xiao, Tu, Zhang, Liu, and
  Sun}]{zhong2020does}
Haoxi Zhong, Chaojun Xiao, Cunchao Tu, Tianyang Zhang, Zhiyuan Liu, and Maosong
  Sun. 2020.
\newblock How does nlp benefit legal system: A summary of legal artificial
  intelligence.
\newblock In \emph{Proceedings of the 58th Annual Meeting of the Association
  for Computational Linguistics}, pages 5218--5230.

\end{thebibliography}
\end{document}